# Seismic Inverse Modeling Method based on Generative Adversarial Network

Pengfei Xie[1], YanShu Yin[2], JiaGen Hou[1,*], Mei Chen[1] and Lixin Wang[2]

1. College of Geosciences, China University of Petroleum, Beijing, 102249

2. School of Geosciences, Yangtze University, Wuhan, 430100

*Corresponding Author, e-mail: 207872336@qq.com

**Key points:**

The paper proposed method bridge a gap between geophysics data and geological knowledge.

The paper proposed method solves 3-D poststack seismic inversion problem, is more useful to application of production and development.

The paper proposed method generate models with low uncertainty and faster speed.

**Abstract:** Seismic inverse modeling is a common method in reservoir prediction and it plays a vital role in the exploration and development of oil and gas. Conventional seismic inversion method is difficult to combine with complicated and abstract knowledge on geological mode and its uncertainty is difficult to be assessed. The paper proposes an inversion modeling method based on GAN consistent with geology, well logs, seismic data. GAN is a the most promising generation model algorithm that extracts spatial structure and abstract features of training images. The trained GAN can reproduce the models with specific mode. In our test, 1000 models were generated in 1 second. Based on the trained GAN after assessment, the optimal result of models can be calculated through Bayesian inversion frame. Results show that inversion models conform to observation data and have a low uncertainty under the premise of fast generation. This seismic inverse modeling method increases the efficiency and quality of inversion iteration. It is worthy of studying and applying in fusion of seismic data and geological knowledge.

# 1 Introduction

Geological modeling has become a conventional mean to guide production in the middle and late stages of petroleum development and it is the core of reservoir characterization. However, there are complicated underground geology and strong reservoir heterogeneity, resulting in limitations of various modeling methods. In applications, it is difficult to predict subsurface geology by geological statistics. Conventional statistical modeling method simulate realizations with high uncertainty and geologists have to combine seismic data with geostatistical modeling. Seismic measurements have high resolution in lateral (Mukerji T et al., 1997), even though seismic data is noisy and low resolution in vertical. Seismic data play a key role in reducing uncertainty of geological modeling (Jeong C et al., 2017). However, it is nearly impossible to find a linear relationship between well logs and seismic data.

The key of seismic inverse modeling is to simulate facies models that can confirm to conditional well points and seismic data. Many statistical modeling methods have been applied to combine geophysical data and geological modeling methods in real applications. There have been two ways: soft data and Bayesian inversion (Doyen P M et al., 2007; Bosch M et al., 2010). In some geological software, seismic data is transformed to probability maps as soft data and soft data can control geostatistical simulation (Dubrule O et al., 2003; Li FM et al., 2007). In the Bayesian inversion, geostatistical simulation provides models as geologically consistent prior models (Eidsvik J et al., 2004; Lochbuhler T et al., 2015). Rock-physics relations can link the seismic data and reservoir property. The forward synthetic seismic data are compared with obtained seismic data to calculate the likelihood, final solutions are posterior models consistent with geological pattern, well data, and seismic data. In this process, sequential method or multi-point geostatistics (MPS) method is used to generate models as prior models (Haas A et al., 1994; Gonzalez E F et al., 2008). The sequential method (Strebelle S et al., 2000) can reflect the statistical relationship and characterize spatial structure incompletely. The MPS can reproduce patterns of training image (Mariethoz G., 2010; Mariethoz G et al., 2014; Pyrcz, M. J et al., 2012), but the computational time of generation is longer than sequential method, which results in the increasement of each inversion iteration. Therefore, the MPS have no ability to reach the requirements of statistic inversion application.

In the field of image recognition, deep learning algorithm has been extensively applied in extraction of data pattern and reproduction of image features. The practical effect of deep learning algorithm is often better than those of traditional methods. One of the most successful generative model algorithms is the generative adversarial networks (GANs). GAN possesses the incomparable advantages in data mining and mode generation and it can characterize the deep hidden data structures.

GAN has been applied widely in the field of geological modeling since it was proposed by

Goodfellow (Goodfellow, I. J et al., 2014). First of all, the CT images of micro-pores were trained by using GAN, the GAN can reproduce the micro-pore structures and implement conditional modeling through the slice constraints (Mosser L et al., 2017). And then, GAN was applied in parametrization and generation of geological models (Chan S et al., 2017). Subsequently, GAN was applied to inversion of seismic waveform (Mosser L et al., 2019). The inversion results conformed well to the observation points, which proved feasibility of inversion algorithm. Laloy learned the two-dimensional and three-dimensional images (which are called training images in multiple-point geostatistics) by DCGAN and optimized based on GPU parallel computation and sampling algorithm, which increased the modeling speed and decreased uncertainty of the model (Laloy E et al., 2012; Laloy E et al., 2018; Laloy E et al., 2017). Zhang Tuanfeng et al. learned the field geological mode by using GAN and achieve conditioning modeling of virtual conditional points, which verified feasibility of GAN in geological modeling (Zhang T F et al., 2019. Zhu L et al., 2019). Song bridged the gap between Geophsics data and geology with GANs (Song S et al., 2021; Song S et al.,2021). GAN-based modeling method was improved continuously subsequently, including improvement of GAN loss function and algorithm optimization. It becomes an intelligent modeling technique with strong applications gradually.

In this study, the binary categories channel-mud training image were learned based on deep convolutional generative adversarial neural network (DCGAN). Moreover, an inversion study of Markov Chain Monte Carlo (MCMC) sampling was carried out by combining the constraints of production wells in practical research area, which achieved fast generation and uncertainty assessment of the models.

## 2 Methodology and Principles

### 2.1 Principle of GAN

GAN consists of two networks with different functions: generator network and discriminator network. Since convolutional neural network has advantages in extraction of image features, the proposed GAN is composed of convolutional neural networks, including a convolutional neural network and a convolutional neural network. GANs extract and characterize feature maps which are used as the input data through the convolutional operation of filter, thus increasing the depth of data expression. Hidden spatial structure of input data, or known as the spatial features of geological model, can be considered more explicitly. The basic principle of convolution is to get several feature matrixes through convolution, unfold them in rows and connect them into vectors which are transmitted into the fully connected layer. The full connection is similar with BP neural network and each feature map can be viewed as neurons which are arranged in the matrix form.

Input data of each convolutional layer has to be transformed through the Eq.(1):

$$V = conv2(W, X, valid) + b \qquad (1)$$

where the input information is V. conv2() is the convolution operation function, W is a weight matrix and X is an input matrix. The third parameter is the type of convolutional operation and b is the bias.

The output information is gained through the Eq.(2):

$$Y = \varphi(V) \qquad (2)$$

where Y is the vector of network output and $\varphi$ is an activation function, such as Sigmoid, tanh and ReLu. In this study, Leaky ReLu (Glorot X et al., 2011) was applied.

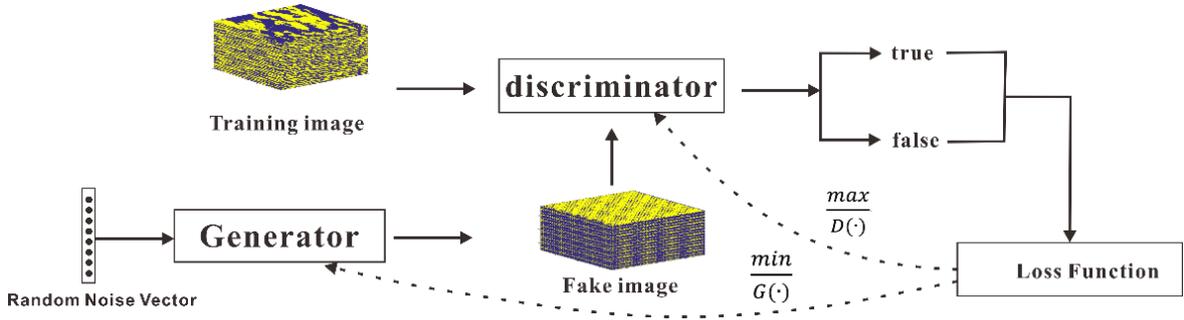

**Fig 1.** Illustration of the GAN structure in context of the geological modeling. The generator gets a random noise vector as latent input and generate a fake image to discriminator. The discriminator distinguishes fake image from real training image and output a "true" or "false" to loss function. Finally, the loss function calculates weight and feedback to generator and discriminator by Eq. (3).

GAN is composed of a generator and discriminator in structures (Fig 1). In the training process, the generator and discriminator reach the equilibrium convergence through a dynamic adversarial process until the generated images are difficult to be distinguished. The generator is a convolutional neutral network and receives one random noise (Z) which represents a series of random vectors that can control distribution of data. This noise vector Z generates a new image (this new image has different distribution with the training image, but similar features). The discriminator is a convolutional network which is used to distinguish the real images from fake ones. It inputs images, and outputs probability of judging reality and feedbacks to the generator. Scanning all data once is called as an epoch and the optimal results are realized after several epochs.

The whole process is an optimization function (Eq. (3)):

$$\min_{G(\cdot)} \max_{D(\cdot)} \{E_{\mathbf{x} \sim P_{data}(\mathbf{X})}[\log(D(\mathbf{X})] + E_{\mathbf{z} \sim p(\mathbf{Z})}[\log(1 - D(G(Z)))]\} \qquad (3)$$

This optimization function can be understood that the Eq. (3) shall be maximized when updating the discriminator (D), but it shall be minimized when updating the generator (G). The total probability is 1. It converges until the probabilities of both discriminator and generator are 0.5. The convergence process is shown in Fig 2.

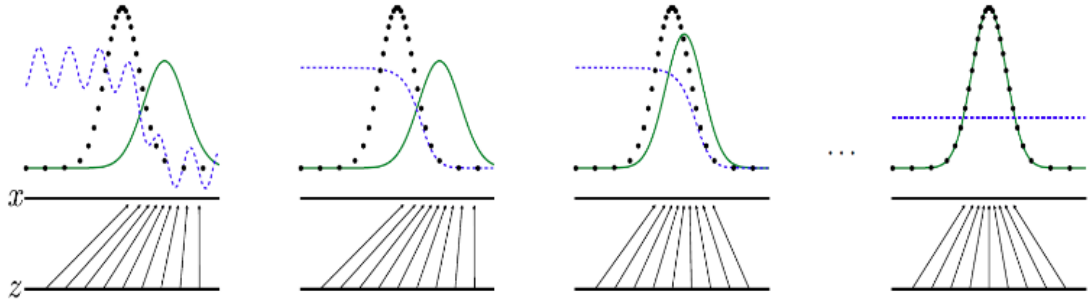

**Fig 2.** GAN converging process (Goodfellow, 2014). In Fig 2, the blue dotted line is the distribution of the discrimination model (D), the black dotted line is the distribution of real data ($p_{data}$), and the green solid line is the distribution ($p_g$) that is learned by the generative model (G). The lower horizontal line is the region of uniform sampling noise (z) and the upper horizontal line is the region of data (x). The upward arrow represents that the random noise is transformed into data (x=G). Fig 2 shows the process of GAN convergence from the left to the right.

**2.2 GAN quality assessment**

It is necessary to assess quality of neural network before Bayesian inversion modeling in order to assure that the training of neural network can extract reservoir spatial structure and statistical features in training images quickly and accurately. The high-quality generative model is an important indicator that the training process of GAN reaches to convergence. The models which are generated by GAN after training have high similarity with the original training image in term of geological features, including continuity and geometrical morphology of channel facies in the model. Therefore, it is necessary to make a quantitative evaluation for training quality of GAN by calculating parameter in spatial statistics. In this paper, the variogram and connectivity functions (CF) of GAN simulations and training image are calculated in different directions (three direction x, y, z) at each lag. The variogram is defined as the variance of the difference between two spatial points of the same facies separated by a given lag distance. while the CF is the probability that there exists a continuous path of the same facies between two points of the same facies separated by a given lag distance. The distribution of spatial statistics curves can certify the stability of simulations.

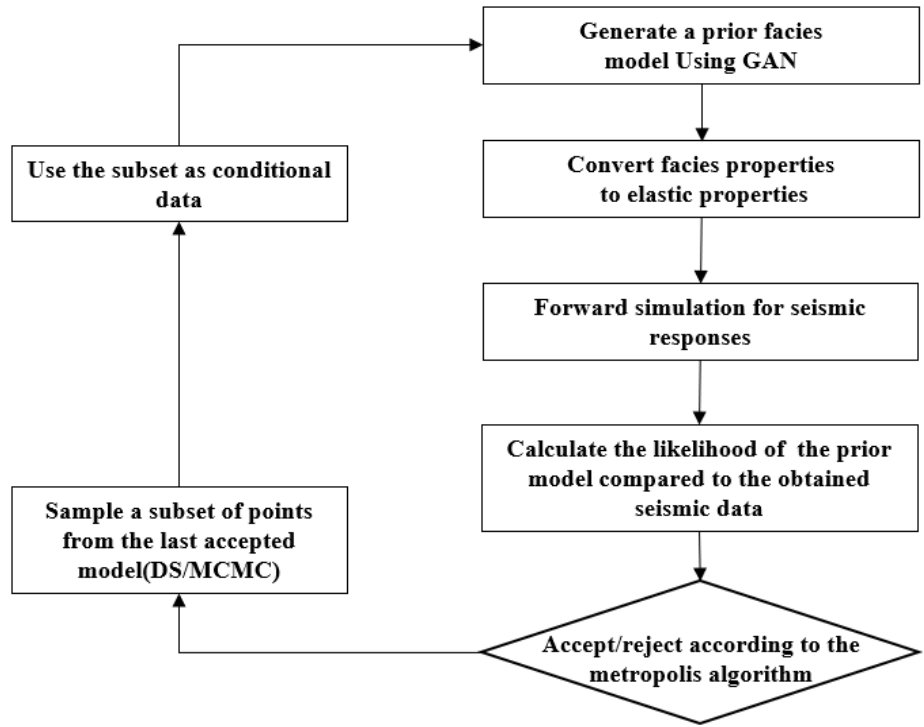

**Fig 3.** A flowchart of Bayesian inverse modeling process with MCMC sampling. In this process, GAN is used as a model generation and it can be applied to seismic inversion. MCMC algorithm samples a subset of points from last accepted model that Markov chain generates. After enough iterations, the facies model generated by GAN can be considered as an inversion result.

**2.3 Bayesian inversion**

The selected GAN after assessment can optimize the learning effect of geological mode and it can recognize and extract the complicated spatial structure and geometric morphology which are difficult to be quantified. It can provide the optimal prior geological information and model generator for the inversion.

In the Bayesian inversion, GAN after training assessment can be used as the generator of geological model. Then the prior facies model is converted to seismic responses in forward simulation. The likelihood of the prior model compared to the seismic data is calculated. The metropolis algorithm samples the posterior probability distribution with the observed data and assumed noise model. Depending on the acceptation/rejection criterion in Markov process, it is possible to obtain a chain of realizations aimed at characterizing the posterior distribution with Metropolis sampling. Since conventional Markov chains can get stuck in a certain state for a long time, DS/MCMC sampling algorithm can help the chains reach the posterior distribution earlier.

This paper use MCMC sampling algorithm to samples a subset of points from the previously accepted model instead of using random selection, and then use the points as conditioning data for next simulated realization. Finally, results that are accepted by Metropolis criterion can be used to calculate the posterior distribution probability (PDF). And we calculate a series of statistical information base on PDF, including expectations, variance, standard deviation (SD) and other statistical features. The inversion result is a multi-dimensional posterior PDF, the mean model represents the most likely result of inversion, the error range is given by the variance model.

The implementation steps of inversion based on GAN are introduced as follows:

(1) After several processes of GAN training and assessment, the optimal GAN is chosen and it can generate the model prior information (priori probability density function). The posteriori probability distribution function (PDF) is solved by the Bayesian formula based on the relationship between the priori information and observation data. The formula is:

$$p(m|d) = \frac{p(d|m)p(m)}{p(d)} \qquad (4)$$

where $d$ is the N-dimensional observation data and $m$ is the M-dimensional model parameter, which can be viewed as a random variable in the inversion process. $p(d)$ is a normalized constant and $p(m)$ is the priori probability density distribution of model parameters, which is known as the data distribution of GAN. $p(m|d)$ is the conditional probability of $m$ to the observation data ($d$), which is known as the posteriori distribution $p(d|m)$ is called as the likelihood function and it is generally written as $L(m)$. The common expression of likelihood function is a multi-dimensional Gaussian distribution:

$$L(m) = \frac{1}{\sqrt{(2\pi)^N |C_d|}} exp\left\{-\frac{(d-d(m))^T C_d^{-1}(d-d(m))}{2}\right\} \qquad (5)$$

where $d(m)$ is the forward modeling response and $C_d$ is the covariance matrix of data. The likelihood function reflects the matching process between the model and data, and the posteriori distribution can be gained through a likelihood function and posteriori probability.

(2) A parallel operation was performed by the MCMC algorithm and several Markov chains are generated by the posteriori distribution. Each chain starts from any point and then is transformed gradually through the transfer probability. Points are adjusted randomly one by one. After enough times of sampling, the global posteriori distribution converges to the stability, and an inversion reservoir model with different standard deviations (theoretically close to the maximum accuracy) is gained.

## 3 Results

### 3.1 Training of GAN

A binary categories channel-mud model was chosen as the training image (Fig 3). This training image was collected from the Australia Morse River valley model in the training image library (http://www.trainingimages.org/training-images-library.html). The total number of grids was 120×150×180 pixels. This training image has evident channel features and is a strip on axis Z. The channel is straight along on axis Y, but it looks like a lentoid with a flat top and convex base perpendicular on axis X. The proportion of facies channel is 0.51 and the proportion of facies mud is 0.49.

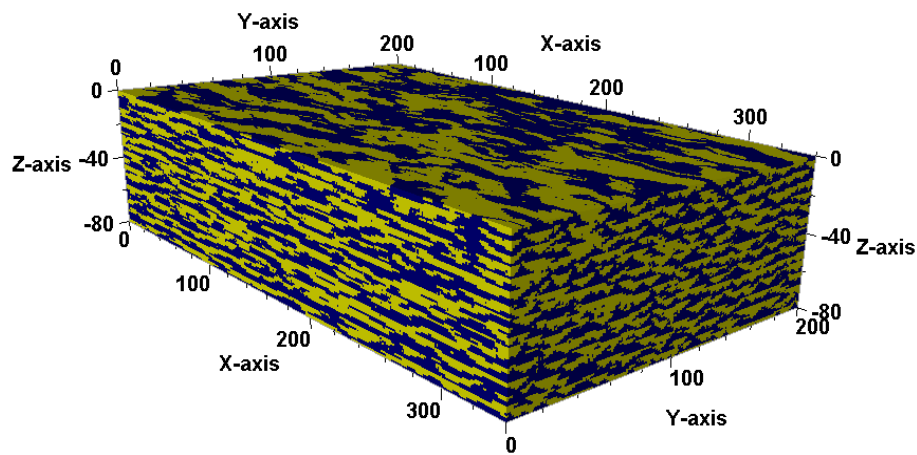

(a)3-D training image

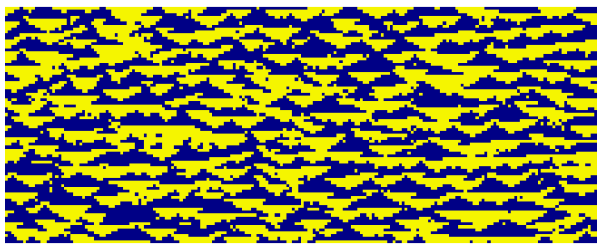

(b)The plane of orientation X

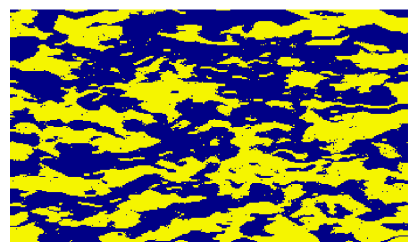

(c)The plane of orientation Z

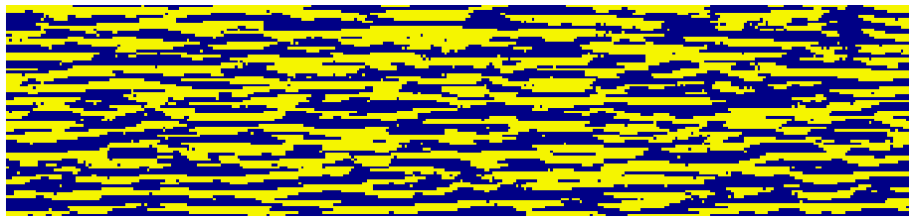

(d) The plane of orientation Y

**Fig 4.** Training image and three planes. On the orientation X, the model is characteristic of a flat top and a convex base. On the orientation Y, the model is a continuous stripe and it presents uneven stripped distributions on the orientation Z. (Yellow represents channel, blue represents mud)

In the GAN algorithm, 3×3×3 kernel size median filter is chosen for feature convolution of the training image. The one-dimensional vectors which can reflect the hidden layer of the space were gained through five convolutional layers and one channel (or called as number of feature maps). In each training period, the batch_size was set 25, indicating that data learning of each training needs 25 batches of processing (batch_size). The generator generates 129×129×129 random model by a 5×5×5 kernel filter through transposed convolution. Under the collaborative operation of intel E3 central processing unit (CPU) and Quadro P5000 graphics processing unit (GPU), GAN finishes 50 epochs in 0.5h and the neural networks after 50 epoches of training are gained. These neural networks can achieve thousands of realizations in 1s.

The random models which are generated by neural networks with 1-16, 17-32 and 33-48 epoches are shown in Fig.4a, b, c. With the increasement of training times, neural networks can learn geological models (e.g. width of rivers on the plane, continuity of lateral channel and morphology of channel of flat top and convex base) in the training images more effectively, and reflect flowing direction, width and continuity features of channel better. It is worth mentioning that training process is not a stable process and training times can affect the training results directly within a certain range. Therefore, the process to the local optimal solution is fluctuating and it generally can only approach to the optimal solution (gradient descent method, details see Arjovsky M et al., 2017). After some times of training, quality of the generative model is still changing. The network training degree is controlled through the W loss function (Arjovsky M et al., 2017). Hence, it is necessary to evaluate the quality of trained neural network. We can evaluate the quantity of realizations generated by neural networks through calculating the spatial function (variogram, connectivity function). While the difference of spatial functions of various facies on different direction is small, the GAN that generates models can be judged as the high-quality neural network.

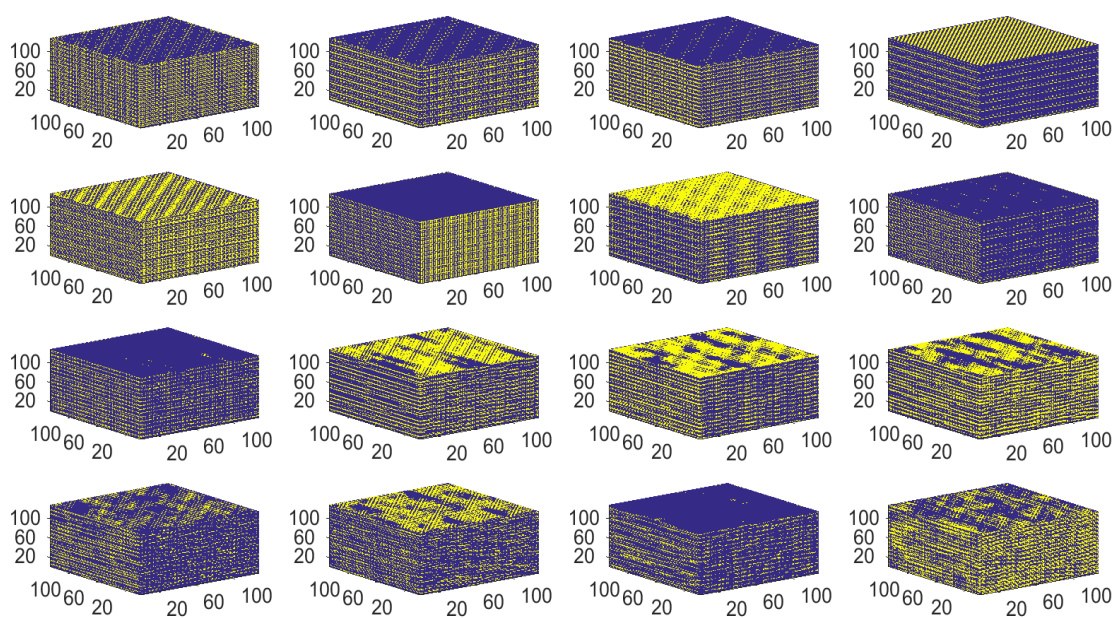

(a) Models derived by GAN after epoch 1-16

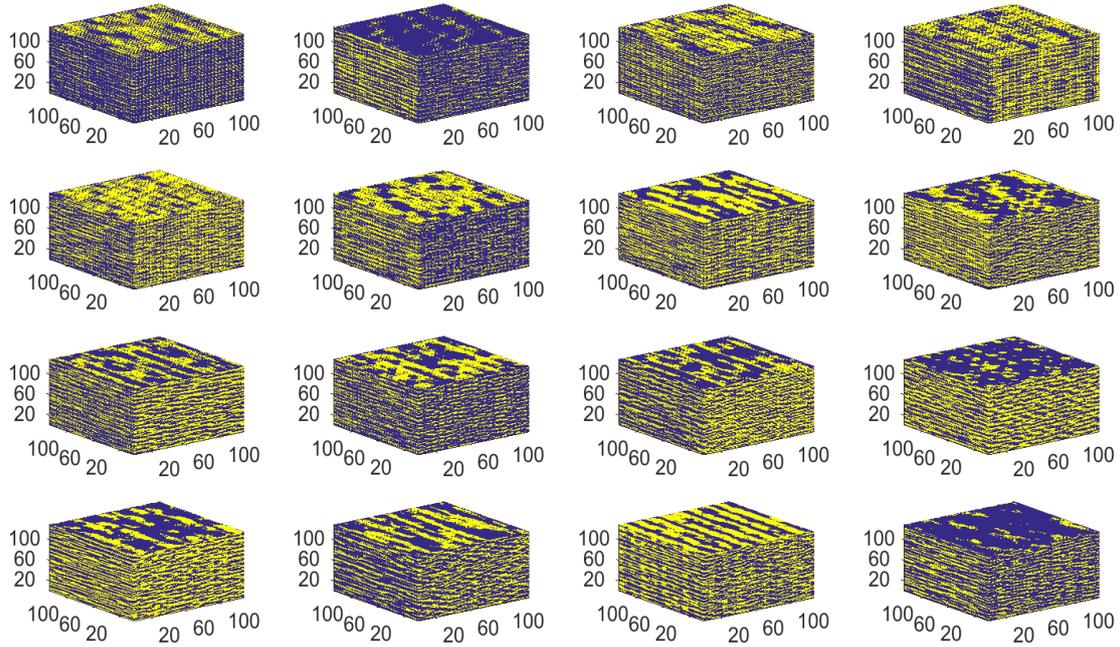

(b) Models derived by GAN after epoch 17-32

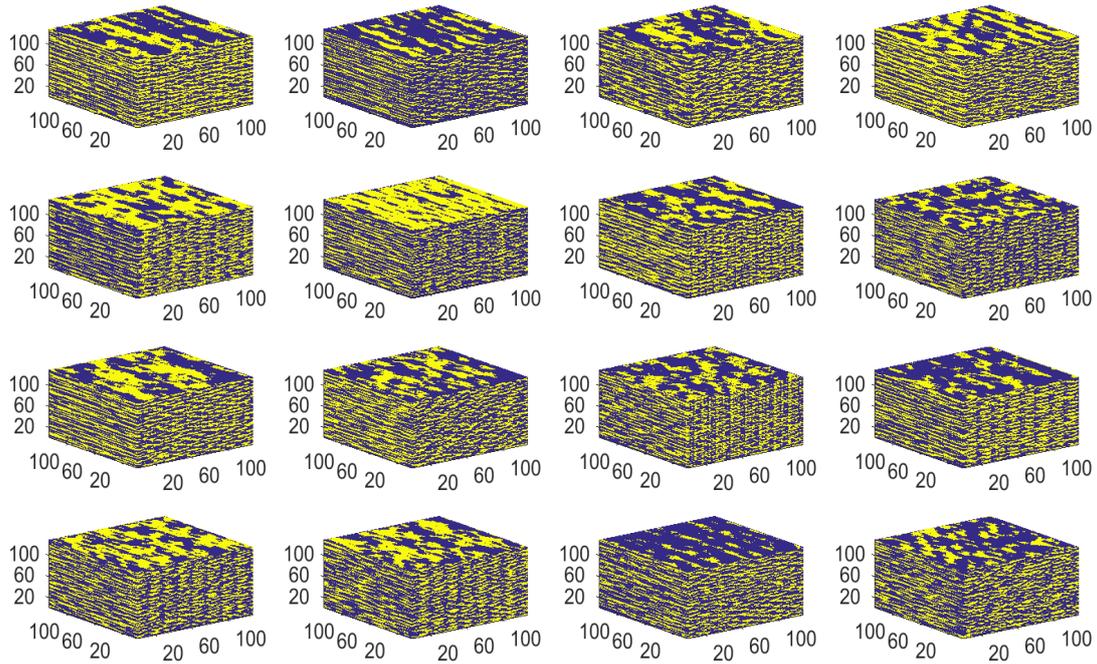

(c) Models derived by GAN after epoch 33-48

**Fig 5.** The models of GAN with 1-48 epochs are generated by GAN in the same random seed. The GAN after higher epoch can generates better quality models until the convergence, these models shows that the result of epoch 46 have the best quality in vision.

The realizations derived by GAN after 46 epochs show similar patterns as TI. Variogram of models derived by GAN after 46 epochs and TI on different directions are shown in Fig 6a, b, c. CF (connectivity function) of models derived by GAN after 46 epochs and TI on different directions are shown in Fig 7a, b, c.

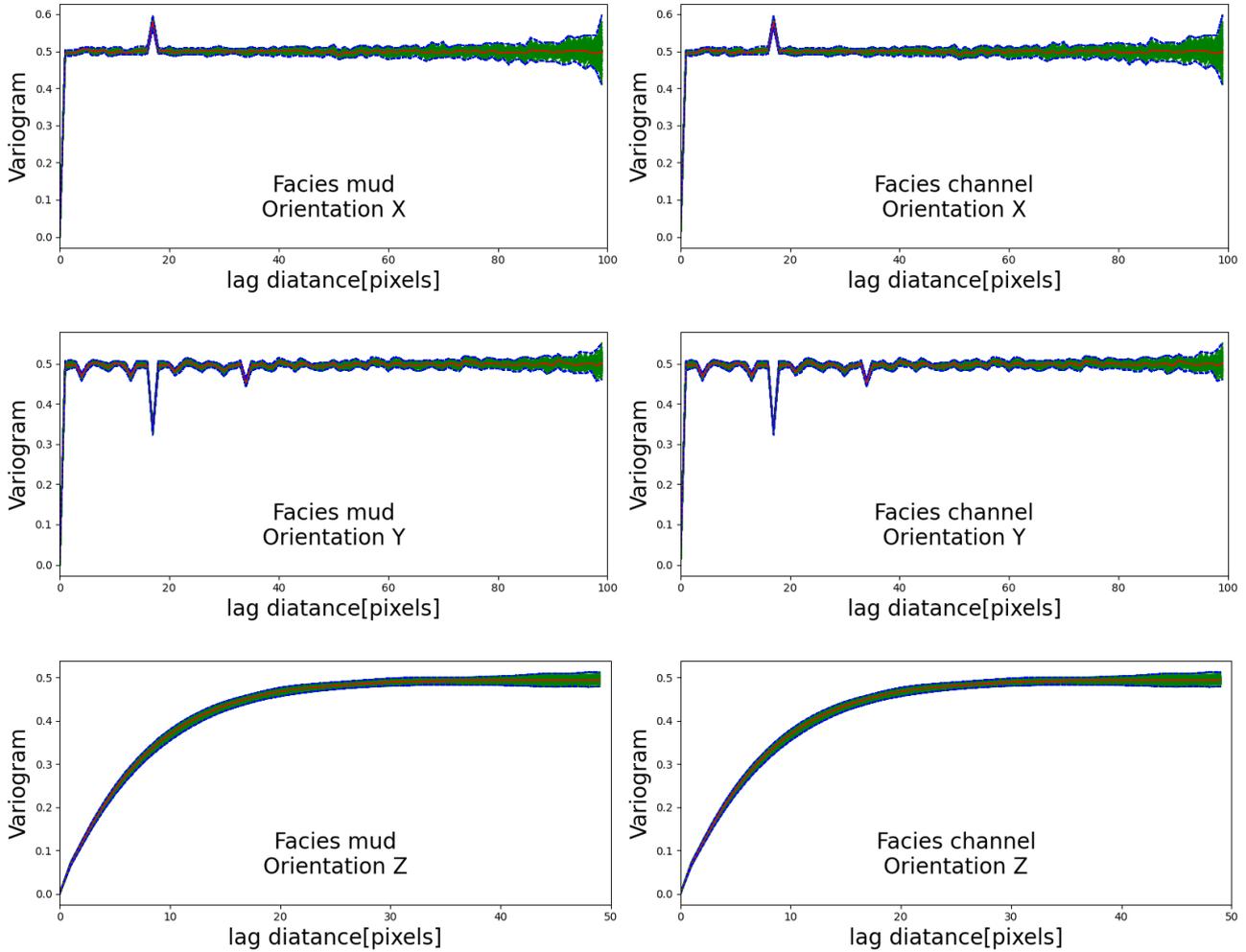

**Fig 6.** Variogram of 100 random models derived by GAN after training epoch 46. The green solid lines represent 100 realizations of size of 129*129*129 GAN realizations. The red solid line signifies the mean for patches of size 100*100*50 randomly selected from the 200*340*80 training image. The two dashed blue lines represent the minimum and maximum value at each lag. The variogram is calculated for each facies in different directions.

The CF and variogram distributions are all calculated for each facies on different directions. The result shows that CF and variogram of models generated by GAN is almost identical to TI. According to definition of variogram, two facies has the same variogram in binary model. So variogram curves of facies mud equal to facies channel. Moreover, two spatial function curves of generations concentrate surrounding the curves of training image. Almost all of models generated by GAN distribute between maximum and minimum of training image (blue dashed lines in fig 6). A good correlation is observed in variograms and CFs between training image and realizations: the maximum absolute deviation of variograms is 0.01, while the maximum absolute deviation of CFs is 0.1. Generations come from training image so that the result is reasonable. Obviously, generations have high stability in statistic. The comparison between generations and training image shows a strong ability in learning features of the training image. Through quality assessment of GAN, the models generated by neural network after 46 epochs (Fig 8) achieves the best effect and the trained GAN can be used in Bayesian inversion.

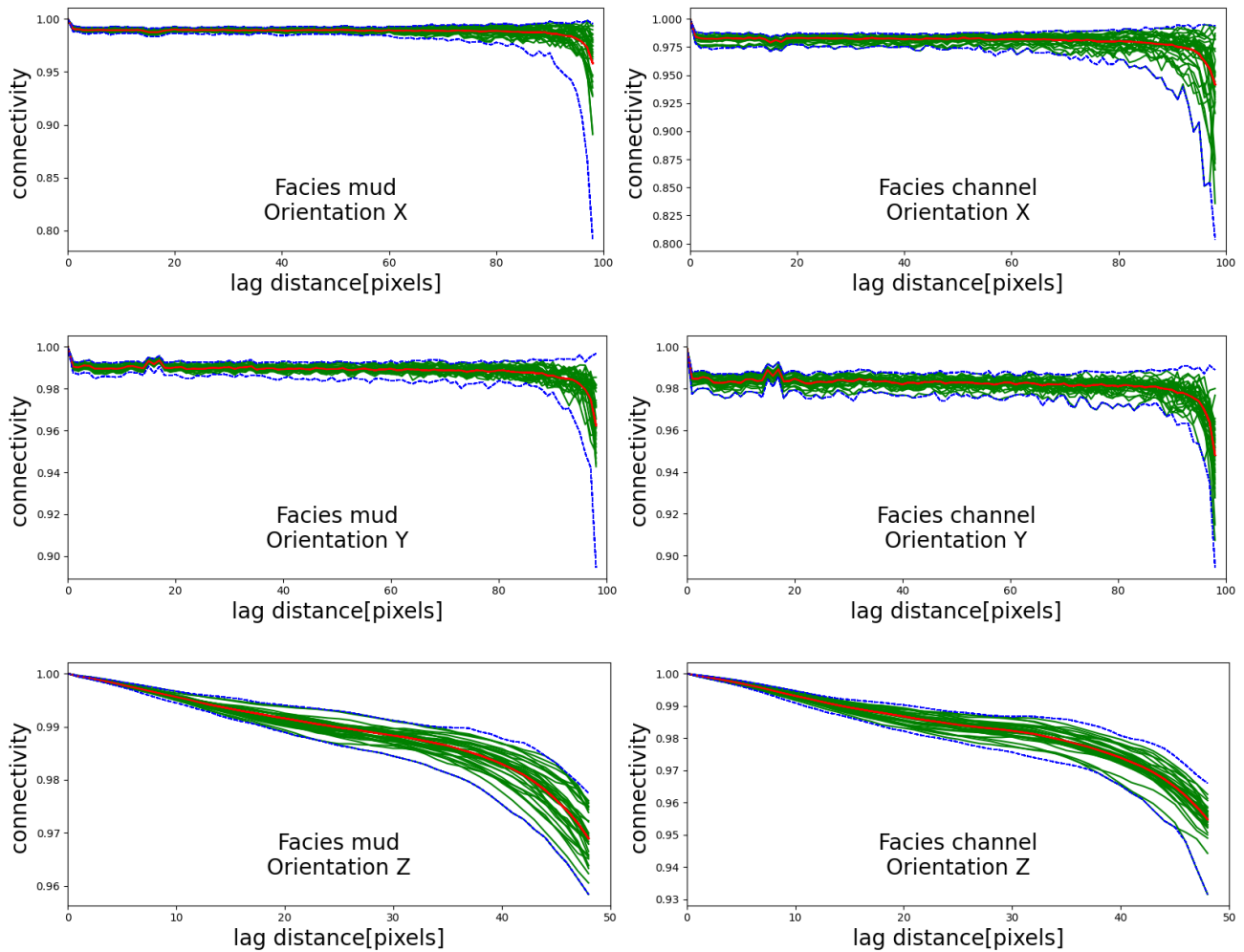

**Fig 7.** Connectivity function (CF) of 30 random models derived by GAN after training epoch 46. The green solid lines represent 30 realizations of size of 129*129*129 GAN realizations. The red solid line signifies the mean for patches of size 100*100*50 randomly selected from the 200*340*80 training image. The two dashed lines represent the minimum and maximum value at each lag. The CF is calculated for each facies in different directions.

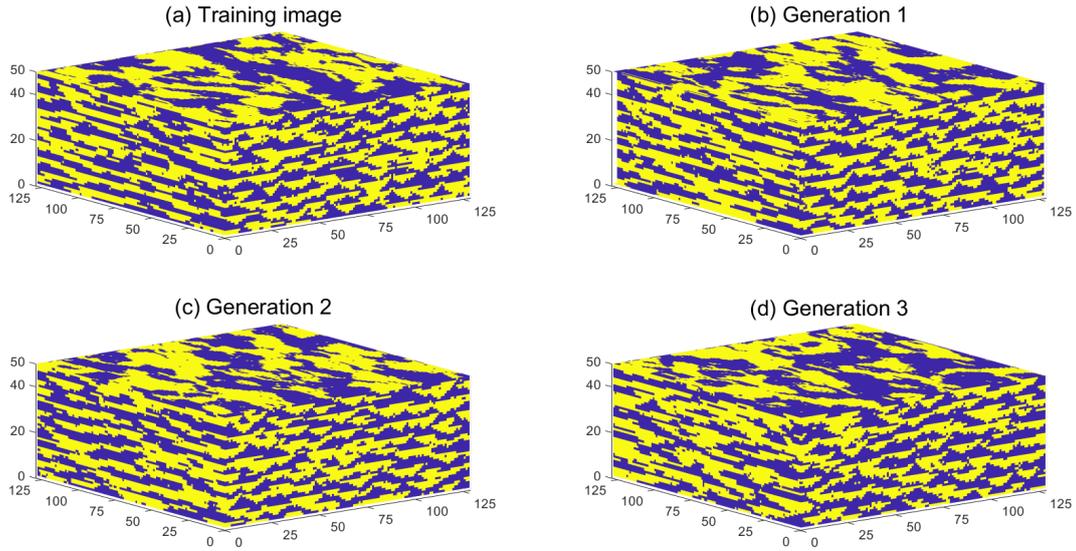

**Fig 8.** (a) training image shown in Figure 4, (b)-(i) randomly selected 129*129*50 realizations derived by GAN of epoch 46. For visual convenience, the original TI was cropped to 129*129*50.

The randomly chosen realizations are generated by GAN after 46 epochs are presented in Fig 8. Fig 6 and fig 7 show the associated variogram and connectivity functions. In vision, the realizations show similar geometric features on plane and vertical profile as the training image. And the facies fraction in TI are same as the average over 100 generations: facies channel: 0.51 versus 0.508 and facies mud: 0.49 versus 0.492.

### 3.2 Inversion modeling based on GAN

The GAN which are chosen by quality assessment can be applied to the Bayesian inversion program. In the case study, a research area with a size of 60×60×30 pixels and a pixel size of 50m×50m×1m was selected, in which 19 wells were set an average interval of 500m. The location of wells is shown in Fig 9-a. The well logs data include velocity, density and lithofacies interpretation result. And channel facies velocity ranges from 4800m/s to 5000 m/s, density ranges from 2.6g/cm$^3$ to 2.8g/cm$^3$, mud facies velocity ranges from 4000 m/s to 4300 m/s, density ranges from 1.9 g/cm$^3$ to 2.4 g/cm$^3$. The elastic property attribute values are distributed in two intervals, we can distinguish channel facies and mud facies by seismic response in forward simulation (40 Hz dominant Ricker wavelet is selected). A seismic data with 40Hz frequency (general frequency) is used to compare with seismic response of the prior model. (Fig 9-b)

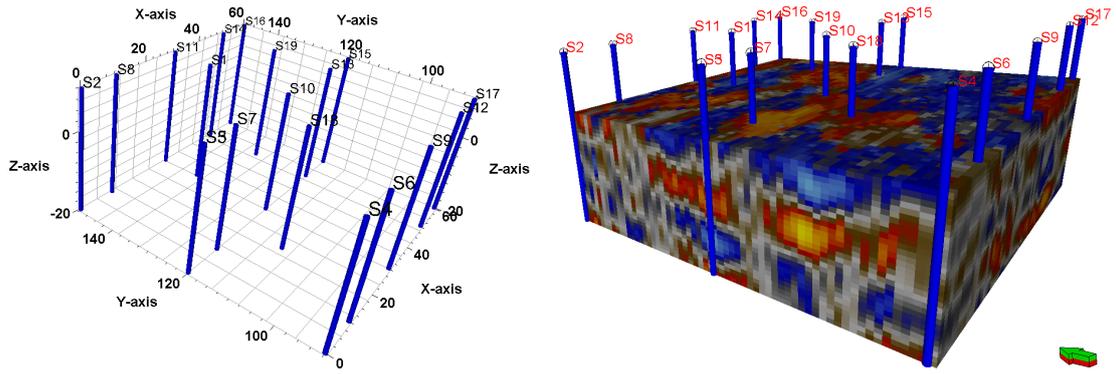

(a) well locations

(b) seismic data

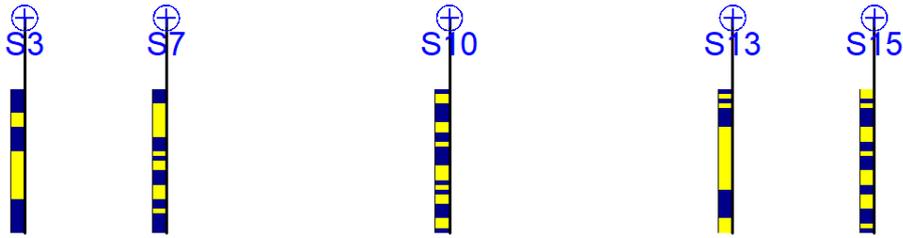

(a) well locations section of S3-S15

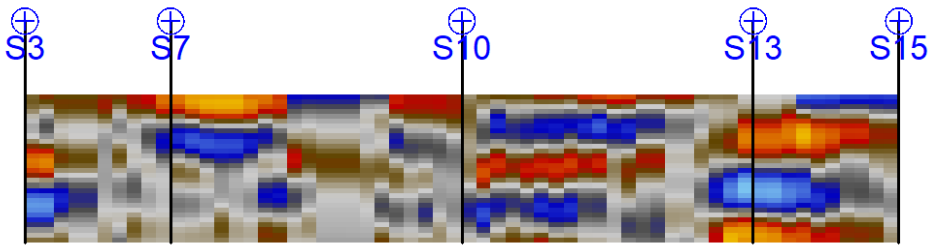

(b) seismic data section of S3-S15

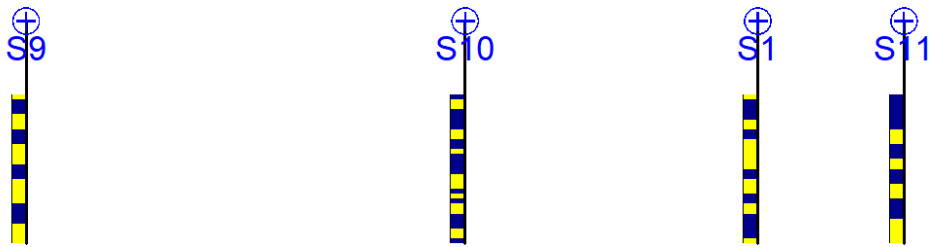

(e) well locations section of S9-S11

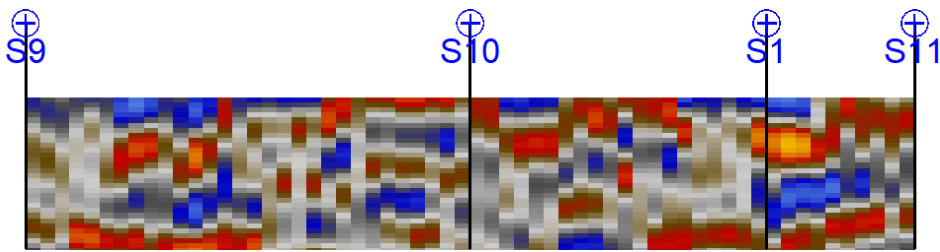

(f) seismic data section of S9-S11

**Fig 9.** (a) Well locations in research area and (b) seismic data with 40Hz. (c-f) well sections and seismic data sections. Facies in wells of section S3-S15 (orientation axes X) are shown in fig9-c, and seismic data response of section S3-

S15 is displayed in fig9-d. Facies in wells of section S9-S11 (orientation axes Y) are shown in fig9-c, and seismic data response of section S3-S15 is displayed in fig9-d.

According to the well data and seismic data, elastic property, we can simulate the response of prior model. As section 2.3 shows, we use the GAN selected by quality assessment to inverse modeling. And GAN generate models with the randomly drawn gaussian noise $(3*3*3)\theta = Z \sim U(-1,1)$ (models show as Fig 10) The prior model of Bayesian framework is chosen as a white noise model (true model), the measurement data have a (standard deviation) SD of 0.01m.

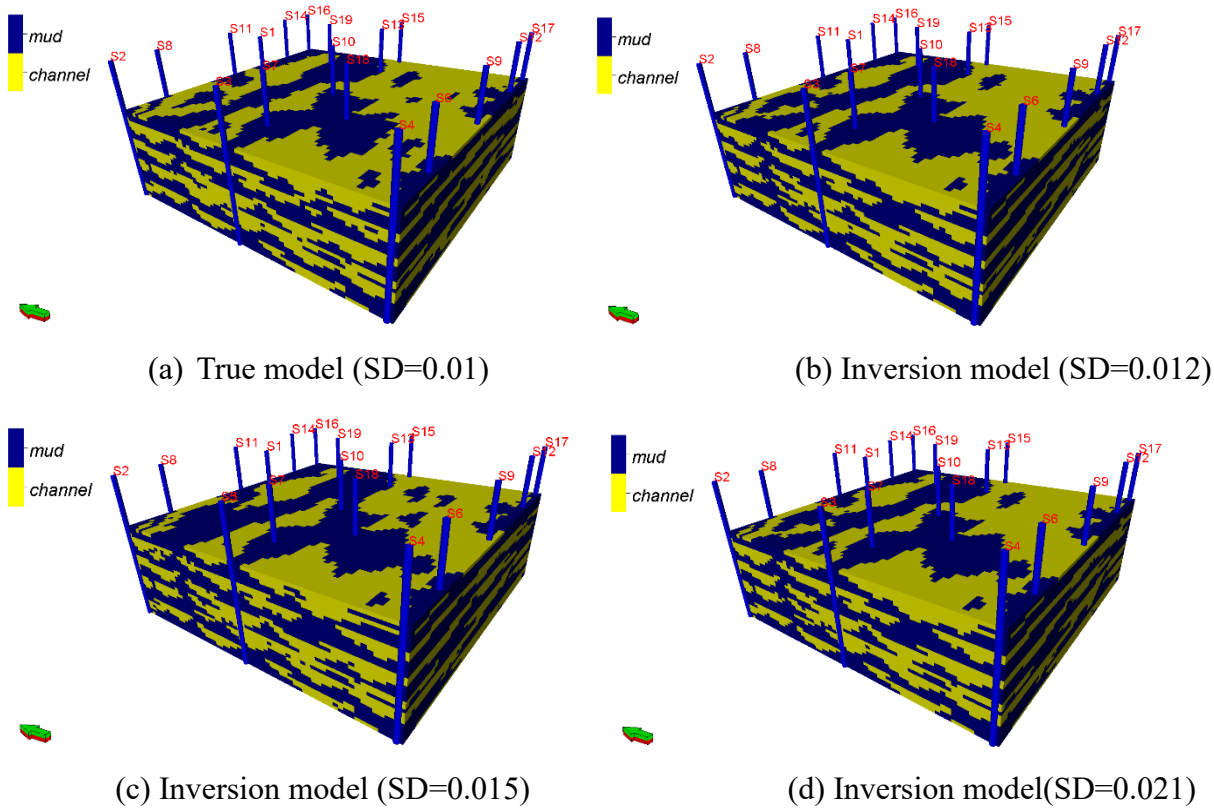

(a) True model (SD=0.01)  (b) Inversion model (SD=0.012)

(c) Inversion model (SD=0.015)  (d) Inversion model(SD=0.021)

**Fig 10**. Lithofacies inversion model with different SD. (a) Ture model (b-d): 3 of 6 Markov chains evolved by MCMC after 30000 iterations per chain. The model dimensions are 60*60*30.

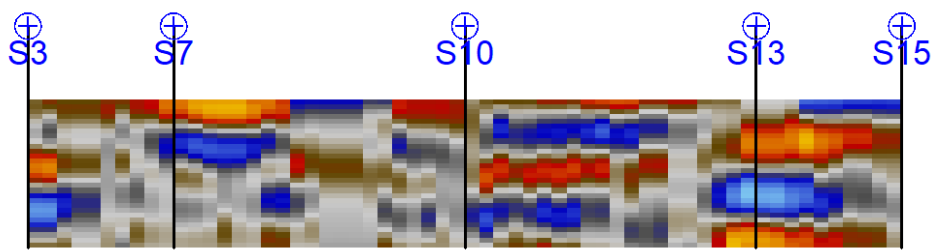

(a) Seismic data in section S3-S15

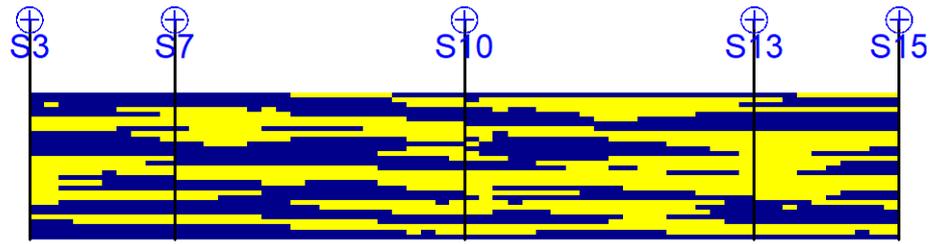

(b) Inversion simulation in section S3-S15

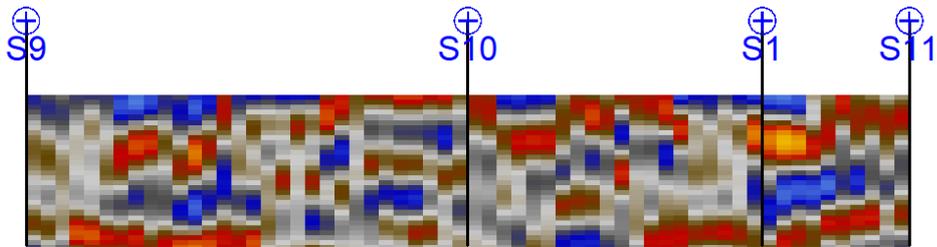

(c) seismic data in section S9-S11

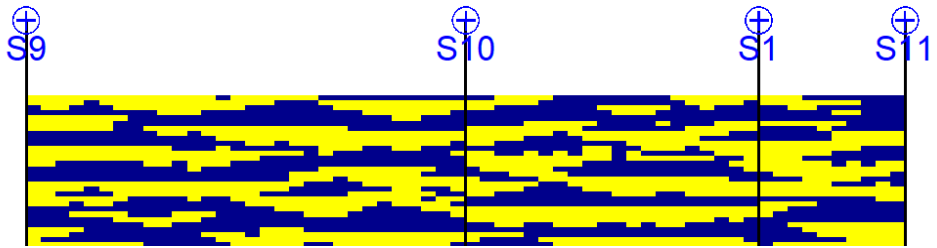

(d) Inversion simulation in section S9-S11

**Fig 11** Comparation from seismic data and inversion result. (a) seismic data of section S3-S15(fig 9-d) and the inversion result is shown in fig11-b. (c) seismic data of section S9-S11(fig 9-f) and the inversion result is shown in fig 11-d. Yellow is facies channel and blue is facies mud.

In inversion framework, Metropolis-Hastings algorithm (an advanced algorithm of Metropolis) in MCMC method was chosen for a parallel calculation in a 12-core CPUs workstation. The 12 Markov chains in parallel generate models using 12 CPUs in a uniform distribution [-1,1] (3*3*3 of θ). The computational parameter is set as 30000 iterations per chain, which cost 1s per iteration, which means 4.1 days on a single core workstation or 8.3 hours on an Intel-Xeon E5 12-core CPUs workstation. The distribution function of each Markov chain is uniform distribution and meets the symmetric random walking. The probability transfer cores are gained from the posteriori distribution, which is known as the probability distribution matrix of facies. In this way, the facies model is acquired. A total of 12×30000 sampling iterations were performed in the whole inversion process, which cost 8h and got 12 results. All of these 12 results conformed to the conditional data completely accompanied with local tiny deviations. Due to the resolution of seismic data, thin channel sands match partially the seismic while thick channel sands match completely the seismic data. (See Fig 11b and 11d) In these results, the minimum standard deviation was 0.013 and the maximum was 0.02. The facies inversion models with different standard deviations are shown in Fig 10b-d (3 results are selected) and the result with the smallest deviation is the optimal solution. Obliviously, the inversion facies models meet well with the priori gaussian white noise model (Fig 10a), showing low uncertainty. This proves that the priori distribution generated by GAN has been used effectively in the inversion process.

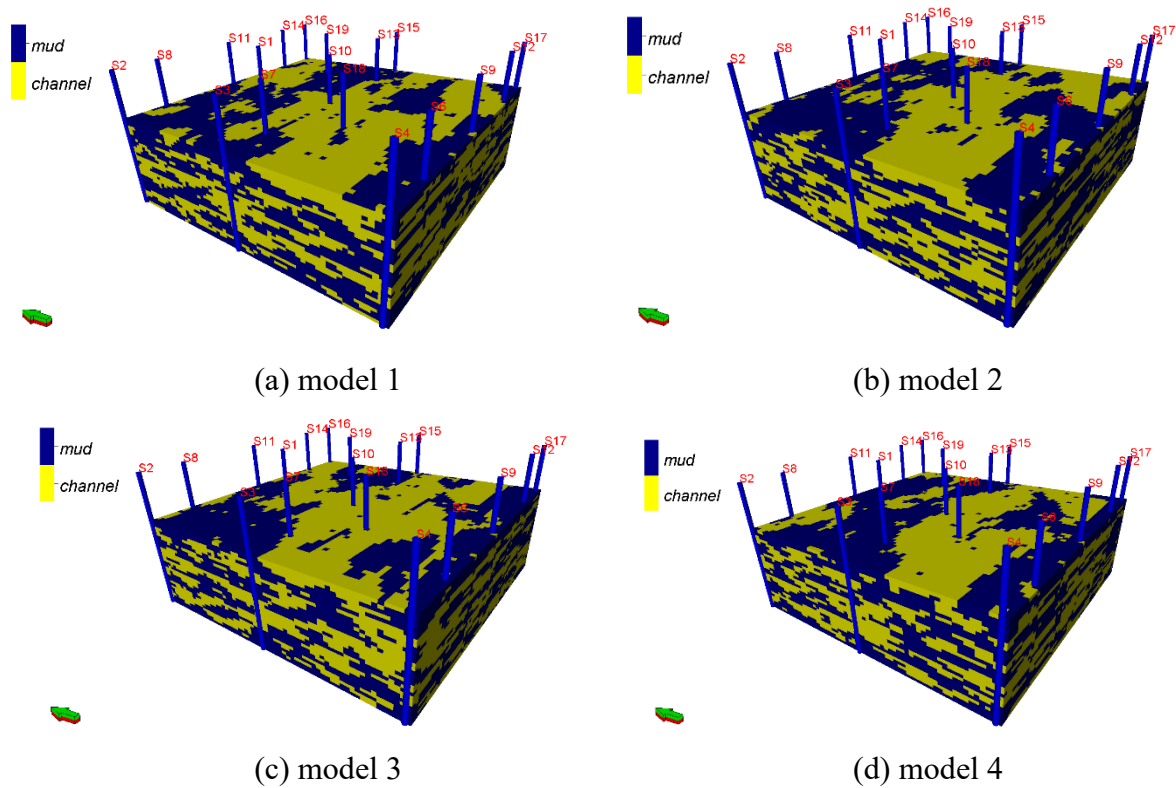

(a) model 1  (b) model 2

(c) model 3  (d) model 4

**Fig12.** Models generated by MPS. The four realizations are generated by software Petrel.

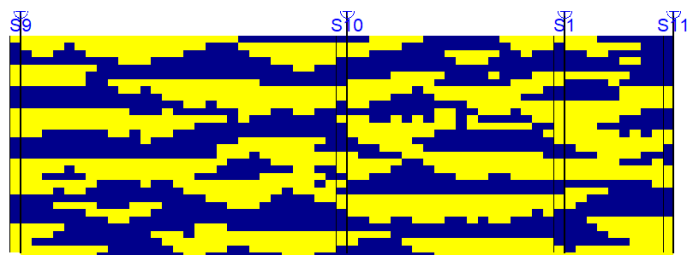

**(a)Inversion in X direction**

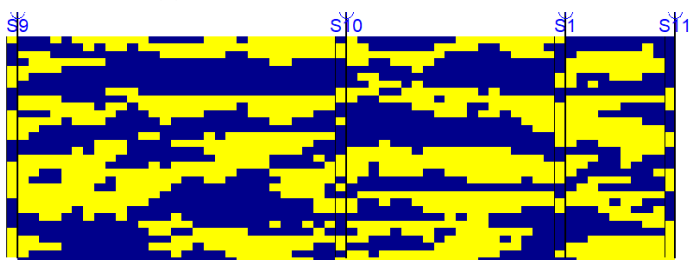

**(b)MPS model in X direction**

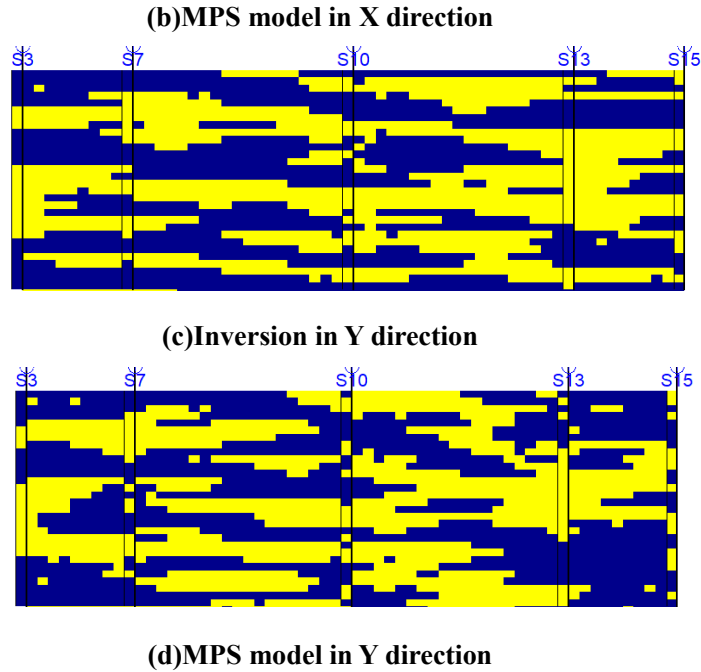

**(c)Inversion in Y direction**

**(d)MPS model in Y direction**

**Fig 13.** Comparison of (a) and (c) the inversion model based on GAN with (b) and (d) the simulation based on MPS. The (a) and (c) vertical sections in orientation X and Y are selected from fig 10b. (b) and (d) vertical sections in orientation X and Y are selected from fig 12a.

The optimal result of inversion model based on GAN and randomly selected simulations generated by multiple-point geostatistics (Strebelle S., 2000) were compared. The four random models which are selected from 100 MPS simulations are shown in Fig 12a-d. In MPS conditional modeling, we use the same training image and create pattern with 10×10×3 search mask. And we simulate models by Snesim based on 19 conditional data (For details, see Strebelle S., 2002), and conditional wells show as Fig 9a). In the comparation of vertical sections between GAN-based inverse modeling and MPS (Fig 13a-d), we can see that the former results have better continuity and stability. In the X direction section, the channel facies of inversion have better geometry than MPS model. In the Y direction section, the channel facies of inversion have better continuity than MPS model, and have less noise. All of results confirm to 19 conditional wells data (100% conditioning accuracy). Given the same conditional data, GAN-based inverse modeling is better. The result shows that the paper proposed approach has an ability to learn geological knowledge and reproduce the pattern of channel in nature.

# 4 Discussions

In this study, the results show that GAN-based inverse modeling method can generate a model with low-uncertainty and high conditioning accuracy. Compared to seismic waveform inversion (Mosser L et al., 2019) and, the inverse modeling method solves three important issues: it uses a single model as a training image to learn pattern of geology, many researchers stress on big data and request a lot of training sets. But the truth is that better data is more important than big data, if a good training image have all patterns we need, and the big data will increase the unexpected computations. The latter point is that our proposed method solves 3-D seismic inversion problem, is more useful to application of production and development. The third one is that MCMC-sampling method and parallel algorithm can solve larger dimensional problem.

Compared to geostatistical inversion using a spatial generative adversarial network (Laloy E et al., 2018), this approach applies geophysics attributes to inverse modeling, generally geologists have to processing the post-stack seismic data for interpretation. This advantage make it become suitable to seismic data interpretation.

There are many solutions in seismic interpretation and stochastic modeling, even though simulations are controlled by conditional data. The inverse modeling can simulate a good quality model with low-uncertainty by sampling method. And researchers often select model subjectively, which is unconvincing. The ability of uncertainty quantification makes the simulation results more convincing. In addition, the inversion has no ability to direct conditioning to wells data, but sampling to conform to conditional data in a target level. It is difficult to achieve absolute match between conditional data and seismic data. If we could expect the seismic interpretation section as conditional data, actually sampling algorithm will be improved (Lixin W et al., 2021). In this process, computational time of inversion require too long to simulate models with large-scale grids. Indeed, genetic algorithm (Shibutani T et al., 2013) suggests that computational time can be reduced. We believe that genetic algorithm can improve our proposed approach. This is a topic worthy of research.

An important issue of our approach is that we need a lot of time to train our network, MPS simulation requires for little time. In limited tests, GAN generates available models with a computational time per model of 0.01s on a single CPU. But GAN reaches convergence for a 180*150*120 binary channel-mud facies model on GPU parallel algorithm after 3 hours training, MPS method created pattern and simulated a model with a computational time per model of 200s on a single CPU (in Petrel software) for the same TI. From the comparison between our approach and MPS, we can see that GAN has a low efficiency of modeling. But the trained GAN can be directly applied to generate models (Song S et al., 2021), which means that GANs have greater strength in a long term. There is no doubt that the training time of GAN is too long to be applied in practical work. Deep learning software and hardware may improve the computational speed, even though this way is limited. Probably establishment of trained GAN database (store all sizes of GAN after training on different sedimentary) is better way to solve the issue.

Another issue is that the convergence of WGAN is not easy to get and the improvements of loss function have been done to optimize the architecture. Just like WGAN-GP (Wasserstein-GAN with gradient penalty), progressive GAN, Info-GAN, recent advances show that WGAN-GP has become the best architecture of GAN. WGAN-GP improves weight clipping of WGAN's critic loss function with gradient penalty. WGAN-GP is the only way to use the same default parameter and train successfully in each architecture (Gulrajani I et al., 2017).

# 5 Conclusions

we proposed a seismic inverse modeling approach that can integrate patterns of geology by GAN algorithm and apply trained GAN to seismic inversion. The results show that realizations conform to seismic data and wells data in the target level, and have a spatial structure of TI. Compared to existing modeling methods, it is able to produce models with lower uncertainty and higher quality. This is a method to bridge the gap between geophysics and geology. Although the 3-5 hours training time and 0.5-1 days inversion time are too long to be applied immediately, the effect is still promising. Main goals of future research will be to improve the quality of inverse modeling, decrease training time and inversion time.

## Acknowledgment

The authors are grateful to 3RG group of Yangtze University. This work was supported by National Science Foundation of China (No.4207020041) and high-performance computer of China University of Petroleum.